\documentclass{article}

\usepackage{microtype}
\usepackage{graphicx}
\usepackage{subcaption}
\usepackage{booktabs} % for professional tables

\usepackage{hyperref}

\usepackage[accepted]{icml2023}

\usepackage{amsmath}
\usepackage{amssymb}
\usepackage{mathtools}
\usepackage{amsthm}

\usepackage[capitalize,noabbrev]{cleveref}

\theoremstyle{plain}

\theoremstyle{definition}

\theoremstyle{remark}

\usepackage[disable,textsize=tiny]{todonotes}
\icmltitlerunning{Unsupervised Adversarial Detection without Extra Model: Training Loss Should Change}

\begin{document}

\twocolumn[
\icmltitle{Unsupervised Adversarial Detection without Extra Model: Training Loss Should Change}



\icmlsetsymbol{equal}{*}

\begin{icmlauthorlist}
\icmlauthor{Chien Cheng Chyou}{NTU}
\icmlauthor{Hung-Ting Su}{NTU,CU}
\icmlauthor{Winston H. Hsu}{NTU,company}
\end{icmlauthorlist}

\icmlaffiliation{NTU}{National Taiwan University}
\icmlaffiliation{company}{Mobile Drive Technology}
\icmlaffiliation{CU}{Columbia University}

\icmlcorrespondingauthor{Chien Cheng Chyou}{eternity6545@gmail.com}
\icmlcorrespondingauthor{Hung-Ting Su}{htsu@cmlab.csie.ntu.edu.tw}
\icmlcorrespondingauthor{Winston H. Hsu}{whsu@ntu.edu.tw}

\icmlkeywords{Machine Learning, Adversarial Detection, Adversarial Attack}

\vskip 0.3in
]



\printAffiliationsAndNotice{} % otherwise use the standard text.

\begin{abstract}

Adversarial robustness poses a critical challenge in the deployment of deep learning models for real-world applications. Traditional approaches to adversarial training and supervised detection rely on prior knowledge of attack types and access to labeled training data, which is often impractical. Existing unsupervised adversarial detection methods identify whether the target model works properly, but they suffer from bad accuracies owing to the use of common cross-entropy training loss, which relies on unnecessary features and strengthens adversarial attacks. We propose new training losses to reduce useless features and the corresponding detection method without prior knowledge of adversarial attacks. The detection rate (true positive rate) against all given white-box attacks is above 93.9\% except for attacks without limits (DF($\infty$)), while the false positive rate is barely 2.5\%. The proposed method works well in all tested attack types and the false positive rates are even better than the methods good at certain types. The source code is available on \href{https://github.com/CycleBooster/Unsupervised-adversarial-detection-without-extra-model}{\textbf{\textit{this GitHub link}}}.
\end{abstract}

\section{Introduction}
\label{introduction}
% The models trained by gradient descent are all vulnerable to adversarial attacks, the attack based on the gradient update on the input \cite{goodfellow2014explaining}. Given some objective for attacks, some unnoticeable perturbations for model's inputs are crafted by gradient descent and added to the inputs, and these changed inputs can totally mislead the targeted model's output. The existing methods still require different parameter customization to defend against different adversarial attacks, but the proposed method can work against different attacks in the same parameters.

% There are two work lines to fight against adversarial attacks, defense or detection. The popular way of defense is adversarial training, the training that makes the model more resistant to be misled. During adversarial training, the inputs will be applied adversarial attacks first and used as training data. This idea works but the effect is still unacceptable. Adversarial training always time-consuming because the attacked should be generated during the training. The next problem is that the model fails to really counteract against adversarial attacks. The last problem is the accuracy drop of the model's original task \cite{goodfellow2014explaining, madry2017towards, raghunathan2019adversarial, zhang2019theoretically}.

Adversarial robustness, a model's ability to withstand adversarial attacks, is a critical issue for deploying deep learning models to real-world applications \cite{carlini2019evaluating}. Adversarial training can strengthen adversarial robustness by using adversarial samples in training, but the side effects are still unbearable, such as significantly increased training time and the accuracy drop of the model \cite{goodfellow2014explaining, madry2017towards, raghunathan2019adversarial, zhang2019theoretically}. What's more, adversarial training only works well for known attack types, which is not practical.

\begin{figure*}[ht]
\vskip 0.2in
\begin{center}
\centerline{\includegraphics[width=\textwidth]{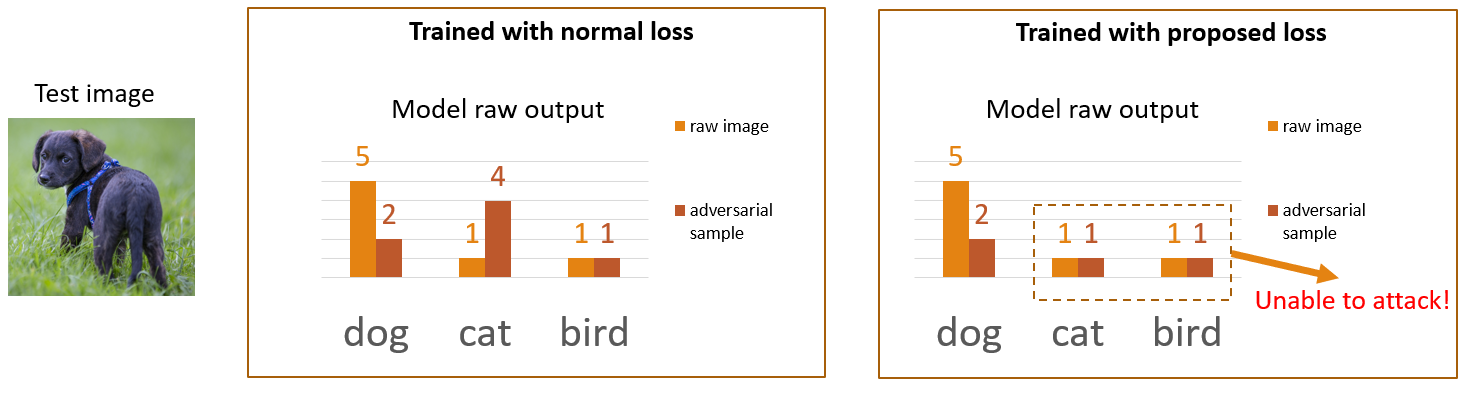}}
\caption{The effect of the proposed training loss and how to use it to detect adversarial samples. Because only the corresponding (dog in this case) raw output can be attacked and other outputs keep the same, the corresponding output becomes a signal to detect adversarial attacks. If no raw output is high, it means that the true output has been attacked.}
\label{fig:how proposed method works}
\end{center}
\vskip -0.2in
\end{figure*}

Another work line is adversarial detection (AD), which attempts to detect adversarial samples before the misled results are adopted. AD methods can be roughly separated into supervised and unsupervised methods \cite{aldahdooh2022adversarial}. In supervised methods, it's possible to reach a satisfying accuracy for detecting adversarial samples. However, they can only detect the attacks they have already known, not to mention the extra training for classifying the given attacked samples \cite{aldahdooh2022adversarial}. The state-of-the-art method now is LiBRe \yrcite{deng2021libre}, which reaches overwhelming detection accuracy and conquers most defects of other supervised methods. The critical problem is that what kind of attacks will come is unpredictable, and including all kinds of attacks is unrealistic.

We choose unsupervised AD as our research direction because of the unknown attack types in advance. However, current unsupervised methods are still very restrictive. Most unsupervised methods need extra models and rely on the distribution of features in the target model or even tune parameters differently based on the datasets or the attack types. Otherwise, they suffer from bad accuracy because the distribution of features changes with datasets and attack types \cite{aldahdooh2022adversarial}. These methods are unaware of the redundant features learned from cross entropy. Therefore, we directly adjust how the target model learns the original classification task.

\begin{figure}[ht]
\vskip 0.2in
\begin{center}
\centerline{\includegraphics[width=\columnwidth]{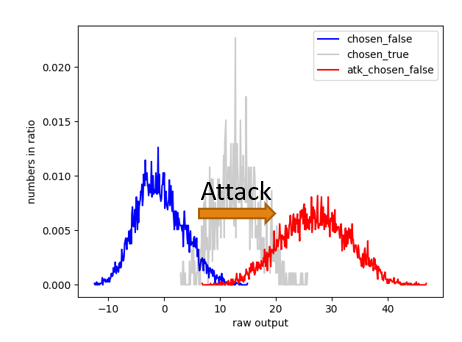}}
\caption{The histogram comparison of attacking class 0's false raw output, the output before softmax. This figure shows that even false raw output can be attacked easily and confirm that there are features for false outputs. Features for false outputs are unnecessary for humans and just give more clues for adversarial attacks. Therefore, this work wants to remove features for false output. PGD attack is applied to the false raw output of class 0 with negative L1 loss (pulling raw output up).}
\label{fig:attack chosen output}
\end{center}
\vskip -0.2in
\end{figure}

Different from the regular AD method, we make part of the target model's output unable to be attacked and detect adversarial samples by the attackable outputs, shown in \ref{fig:how proposed method works}. The key idea is that adversarial samples are a combination of the feature learned from the training dataset \cite{ilyas2019adversarial}. The features for part of the model's outputs, which means these features can affect the corresponding outputs, are useless and can be removed, and then the corresponding outputs become unable to be affected by adversarial attacks. The model's raw outputs, the output before softmax, can be separated into two parts, the true outputs and the  false outputs. The true outputs belong to the corresponding class to the input's class, and the false outputs do not. We find that all model's raw outputs can be attacked with only L1 loss, like figure \ref{fig:attack chosen output}. That is, there are features for false outputs. Considering cross-entropy loss presses down false outputs, the feature for false output could be “why they are not something”, but this is not a good way for humans to classify objects' categories. Human considers “why they are something”, and add points to the corresponding class. Therefore, the feature for false outputs should be unnecessary and just give adversarial attacks more clues to attack the target model. This work wants to remove the features for false outputs by changing training loss, so false outputs become adversarially robust and only true outputs can be attacked. The true outputs' value is lower when attacks happen, so they become a signal for adversarial attacks.

\textbf{We assume that directly keeping all the false raw outputs the same can remove the features for false outputs}. The objective is achieved by forcing all the false raw outputs in a mini-batch to uniform distribution during training. The details will be explained later in section \ref{details for false output's loss}. The assumption works: the false raw output will be very hard to attack, and only the true output can be attacked. What's more, the range of false raw outputs is very separated from true raw outputs. Based on this property, we design a straightforward method for AD. If the maximum raw output is lower than the threshold or the minimum raw output is lower than the other threshold, the input should have been attacked already. There is no complicated algorithm, no extra model, and very low training overhead. The following are our contributions:
\begin{itemize}
\item We confirm that the behavior of cross entropy loss gives more features for adversarial attacks, and features for false outputs are unnecessary.
\item We proposed a new training loss set that can effectively protect the false outputs from many kinds of white box adversarial attacks.
\item We proposed a very simple but effective AD method based on the proposed training loss. Only the target model is used, and there is no extra model. The overhead of supervised and unsupervised detection methods no longer exists, such as extra training, extra computing, extra model weights, hyperparameter tuning, or trade-off for accuracy. 
\item In addition, the true positive rates are above 93.9\% with a false positive rate of merely 2.5\% against all given white-box attacks except for the attack without constraint $\epsilon$. All the attack types are unknown for the target model.
\end{itemize}

\section{Related Work}

Adversarial attacks aim to mislead the target model and force the target model gives the result they want. Depending on what model knowledge is known in advance, adversarial attacks can be categorized into three types, white box attacks, gray box attacks, and black box attacks \cite{aldahdooh2022adversarial}. White box attacks mean everything about the target model is known, including the target model's structure, the training data, the weights, its inputs, and outputs. These attacks can directly compute gradient descent for some objective to get the input perturbations they want \cite{goodfellow2014explaining, kurakin2018adversarial, moosavi2016deepfool, madry2017towards}. Gray box (or semi-white box) attacks and black box attacks are not discussed in this paper. The strength of attacks is evaluated based on some hyperparameter: $\epsilon$ for the norm bound of attacks, how many iterations to update attacks, and how long a step goes.

AD counts on finding the special feature of adversarial samples. Supervised methods are a direct and effective direction. Statistics and patterns of the features can be found in adversarial samples, but they are not general to all kinds of attacks. On the other hand, the uncertainty methods seem to work better because they can deal with all kinds of attacks at the same time, just like LiBRe  \yrcite{deng2021libre}. In this work, a lightweight bayesian neural network (BNN) is connected to a pretrain model, a classification model or even an object detection model, and refine the BNN model with given adversarial data from many types of adversarial attacks. This method solves the computing and training problem of BNN and reaches a spectacular accuracy for detecting the given kinds of adversarial attacks.

However, the type of adversarial attacks is unknown in the wild. That's why unsupervised methods should be applied in the real scenario. Unsupervised methods need to detect adversarial samples with only the original training data. Because the adversarial samples are unknown, the detection threshold is chosen based on the given false positive rate (FPR) for the clean testing data. There are too many different methods, so we only introduce the comparison methods in the experiments. Feature squeezing (FS) and MagNet are based on the property that if the input is attacked, the change in the input will change the output a lot \cite{xu2017feature,meng2017magnet}.  Deep neural rejection (DNR), Selective and feature based adversarial detection (SFAD) and Neural-network invariant checking (NIC) use the features in the target model to check whether the target model works properly \cite{sotgiu2020deep,aldahdooh2023revisiting,ma2019nic}. In these unsupervised methods, the overhead is big. The hyperparameters are hard to tune and most need extra training and extra model, but their accuracies are still infeasible \cite{aldahdooh2022adversarial}.

\section{The Proposed Training Loss and Adversarial Detection}

We find that there are features for false outputs shown in figure \ref{fig:attack chosen output}, and the goal is to reduce false output's features by constraining false raw outputs around the same value. In this way, only true outputs can be attacked and become the signals for adversarial attacks. True raw outputs will be high without attacks, and be low when attacks happen. The following will introduce the preliminary in section \ref{preliminary}, and then explain how to reduce features for false outputs by the loss for false outputs in section \ref{details for false output's loss}. The training method is further improved by replacing the cross-entropy loss with the loss for true outputs in section \ref{details for true output's loss} and removing the model's trend of one-hot output in section \ref{remove one-hot trend}. After applying the proposed training loss, only the true output can be attacked. This property can be used to detect adversarial attacks and the method is introduced in section \ref{proposed adversarial detection}.

\subsection{Preliminary}
\label{preliminary}
Kullback–Leibler divergence (KL divergence) will be the main subject of discussion. The objective of KL divergence is to measure the distance from a source probability distribution, usually the softmax outputs of a deep learning model, to a destination probability distribution. Cross entropy (CE) gets a different meaning in information theory, but CE's gradient is the same as KL divergence's. In most classification tasks, only one class is true, also called one-hot target.

When computing gradients of a deep learning model, the raw output vector $\boldsymbol{Y}$ from a sample after the last layer of the model is always followed with the softmax function $\sigma$ to get the probability output vector $\boldsymbol{Y'}$:
\begin{equation}\label{softmax}
\boldsymbol{Y'}=\boldsymbol{\sigma}(\boldsymbol{Y})
=\frac{e^{\boldsymbol{Y}}}{\sum_{y\in\boldsymbol{Y}}e^{y}}
\end{equation}
Given the softmax output $y'$ of a class and the label value $l$, the gradient of CE corresponding to the raw output $y$ is very simple.
\begin{equation}\label{gradient of CE}
\frac{\partial{CE}}{\partial{y}}=y'-l
\end{equation}
The gradient of raw output will be used to derive the proposed training loss in section \ref{details for true output's loss}.

\subsection{Reduce Features for False Outputs}
\label{details for false output's loss}
Constraining the false output around a certain value is hard because the suitable value is unknown. Applying KL divergence to learn a uniform distribution can constrain false outputs to a suitable value automatically.  We apply KL divergence to false outputs $f$ from set $F$, which contains \textbf{all false outputs from all samples in a mini-batch}, and constrain them to uniform distribution $U$ of false outputs.
\begin{equation}
Loss_{false\_output}=-\sum_{f\in\boldsymbol{F}}{U(f)\log\frac{\sigma_{f}(\boldsymbol{Y_F})}{U(f)}}
\end{equation}
After applying $Loss_{false\_output}$, the range of false raw outputs becomes very narrow decided by the model itself automatically. The most important is that false raw outputs are hard to attack with negative L1 loss now.

\subsection{Replace the Cross-entropy Loss}
\label{details for true output's loss}
We find that pulling true outputs for each class through a mini-batch gets better results than the normal cross-entropy loss. The reason is that when some classes are less accurate, their true raw outputs prefer to be lower, and false raw outputs prefer to be higher. When applying CE, the softmax value of accurate classes and inaccurate classes are similar, so the gradients are also similar. Therefore, CE keeps training accurate classes while they are good enough, and inaccurate classes still fail to get higher gradients. The result is that true raw outputs from inaccurate classes are much lower. If true and false output values are mixed without attacks, we fail to identify the normal situation. The comparison is shown in section \ref{MHKL vs CE}. 

When pulling true outputs in each class, there are several true outputs and the number of them can be different in a mini-batch. The idea to get equivalent loss with CE is that given all false outputs are the same value, the true raw outputs with the same value should get the same gradient. Based on the equation \ref{gradient of CE}, a raw output $y$ should correspond to the same softmax output $y'$ to get the same gradient $y'-l$.

At first, considering one true raw output with the value $y_t$ and $n$ number of false raw outputs with the same value $y_f$, we want to get the relation between $y$ and $y'$ for the true output. The softmax equation \ref{softmax} can change to:
\begin{equation}
\begin{split}
y'_t&=\frac{e^{y_t}}{\sum_{y\in\boldsymbol{Y}}e^{y}}
=\frac{e^{y_t}}{e^{y_t}+\sum_{f\in\boldsymbol{F}}e^{y_f}}
=\frac{e^{y_t}}{e^{y_t}+n\cdot e^{y_f}}\\
&=\frac{e^{y_t-y_f}}{e^{y_t-y_f}+n}
\end{split}
\end{equation}
Then, given a different false output's number $n'$ in a different mini-batch, if we still want to get the same $y'_t$ with the same $y_t$ and $y_f$, we can just add a bias $b$ to the true output:
\begin{equation}
\begin{split}
\because y'_t&=\frac{e^{(y_t+b)-y_f}}{e^{(y_t+b)-y_f}+n'}=\frac{e^{(y_t+b-\log{n'}+\log{n})-y_f}}{e^{(y_t+b-\log{n'}+\log{n})-y_f}+n'\cdot\frac{n}{n'}}\\
&=\frac{e^{y_t-y_f}}{e^{y_t-y_f}+n}\\
\therefore b&=\log{n'}-\log{n}
\end{split}
\end{equation}
The relation between $y$ and $y'$ for the false output is changed but the gradient sum of false outputs is the same.
\begin{equation}
\sum_f{\frac{\partial{KL}}{\partial{y_f}}}=\sum_f{y'_f}=1-y'_t
\end{equation}
In short, given one true output and $n'$ false outputs, to obtain the equivalent loss to $n$ false output, we just add a bias \(b=\log{n'}-\log{n}\) to the true output.

On the other hand, the number of true outputs for a class in a mini-batch can change. The target now is to get the same relation between $y$ and $y'-l$ when the number of true outputs is different. Considering that there are $n_t$ true outputs with the same value $y_t$ and $n_f$ false outputs with the same value $y_f$, and we still get the relation between $y$ and $y'$. The label of multi-hot objective is $1/n_t$ for true outputs:
\begin{equation}
\begin{split}
 y'_t-1/n_t
 &=\frac{e^{y_t}}{\sum_{t\in T}e^{y_t}+\sum_{f\in F}e^{y_f}}-1/n_t\\
 &=\frac{e^{y_t}}{n_t\cdot e^{y_t}+n_f\cdot e^{y_f}}-1/n_t\\
 &=\frac{1}{n_t}(\frac{e^{y_t}}{e^{y_t}+\frac{n_f}{n_t}\cdot e^{y_f}}-1)
 \end{split}
\end{equation}
We can find the gradient is equivalent to one-hot objective averaged by the number of true outputs $n_t$, and $\frac{n_f}{n_t}$ is the averaged number of false outputs for each true output. That is, the gradient of multi-hot KL divergence with $n_t$ true outputs and $n_f$ false outputs equals to the gradient of the expected value of one-hot CE with 1 true output and $\frac{n_f}{n_t}$ false outputs. 

\begin{table*}[t]
\caption{The true positive rate (TPR\%) and false positive rate (FPR\%) of different unsupervised AD methods against well-known white box attacks on CIFAR-10.}
\label{unsupervised vs white box attacks table}
\vskip 0.15in
\begin{center}
\begin{small}
\begin{sc}
\begin{tabular}{lcccccccccccc}
\toprule
  & \multicolumn{2}{c}{FS} & \multicolumn{2}{c}{MagNet} & \multicolumn{2}{c}{DNR} &\multicolumn{2}{c}{SFAD} & \multicolumn{2}{c}{NIC} & \multicolumn{2}{c}{\textbf{proposed}} \\
\cmidrule(lr){2-3}\cmidrule(lr){4-5}\cmidrule(lr){6-7}
\cmidrule(lr){8-9}\cmidrule(lr){10-11}\cmidrule(lr){12-13}
Attack type ($\epsilon$)
&TPR$\uparrow$&FPR$\downarrow$&TPR$\uparrow$&FPR$\downarrow$&TPR$\uparrow$&FPR$\downarrow$&TPR$\uparrow$&FPR$\downarrow$&TPR$\uparrow$&FPR$\downarrow$&TPR$\uparrow$&FPR$\downarrow$\\
\midrule
FGSM (8)
&29.33&5.07&0.72  &0.77&32.09&10.01&67.94&10.9&43.64&10.08
&\textbf{100}&\textbf{2.5} \\
FGSM (16)
&35.34&5.07&3.11  &0.77&31.35&10.01&79.9  &10.9&58.48&10.08
&\textbf{100}&\textbf{2.5} \\
BIM (8)
&8.74  &5.07&0.56  &0.77&4.27  &10.01&18.12&10.9&\textbf{99.95}&10.08
&\textbf{93.9}&\textbf{2.5} \\
BIM (16)
&0.34  &5.07&0.69  &0.77&17.07&10.01&45.35&10.9&\textbf{100}   &10.08
&\textbf{95.8}&\textbf{2.5} \\
PGD (8)
&8.2    &5.07&0.57  &0.77&11.34&10.01&29.49&10.9&\textbf{100}   &10.08
&\textbf{95.1}&\textbf{2.5} \\
PGD (16)
&0.2    &5.07&0.66  &0.77&25.11&10.01&52.9  &10.9&\textbf{100}   &10.08
&\textbf{99}&\textbf{2.5} \\
DF ($\infty$)
&39.18&5.07&57.33&0.77&30.2  &10.01&89.57&10.9&\textbf{84.91}&\textbf{10.08}
&78.8&2.5 \\
\bottomrule
\end{tabular}
\end{sc}
\end{small}
\end{center}
\vskip -0.1in
\end{table*}

In summary, we apply KL divergence to all outputs from all samples $x_c$ of a class $c$, and average through each class with the number of classes $N_C$. The destination distribution is a multi-hot distribution $M$. We temporarily add biases vector $\boldsymbol{B_c}$, which depends on the false-true ratio of class $c$ and is zero for false outputs, to the raw output vector $\boldsymbol{Y_c}$. That is:
\begin{equation}
\label{true output's loss}
Loss_{true\_output}=
-\frac{1}{N_C}\sum_c
\sum_{x}{M_{c}(x)\log\frac{\sigma_{x}(\boldsymbol{Y_c}+\boldsymbol{B_c})}{M_{c}(x)}}
\end{equation}
When using $Loss_{true\_output}$, the false outputs from less accurate classes are compared with each other. They are similar to each other, so the punishment becomes lower. Therefore, the raw outputs from  less accurate classes are much higher than applying CE and help to get a lower false positive rate for the proposed AD method.

\subsection{Remove the Trend of One-Hot Output}
\label{remove one-hot trend}
After applying the proposed two losses, we find that the target model prefers to keep one raw output high. When pressing down the maximum raw output with L1 loss by adversarial attacks, another raw output raises. On the other hand, pulling false raw outputs directly becomes much more difficult now.

To remove this trend, the idea is to generate the condition when every raw output should be low and penalize the output with a high value. To achieve this goal, we apply adversarial attacks to part of the training data in the mini-batch. The objective of adversarial attacks is only to press down the true raw output to generate the condition when one-hot output should not exist. For these adversarial samples, only the loss for false outputs is applied to penalize one-hot outputs. The ratio of adversarial samples can be small, 10\% is enough. If the ratio is too high, the training time will become much longer and the accuracy of the original classification task will be lower.

\subsection{The Proposed Adversarial Detection}
\label{proposed adversarial detection}
\begin{figure}[ht]
\vskip 0.2in
\begin{center}
\centerline{\includegraphics[width=\columnwidth]{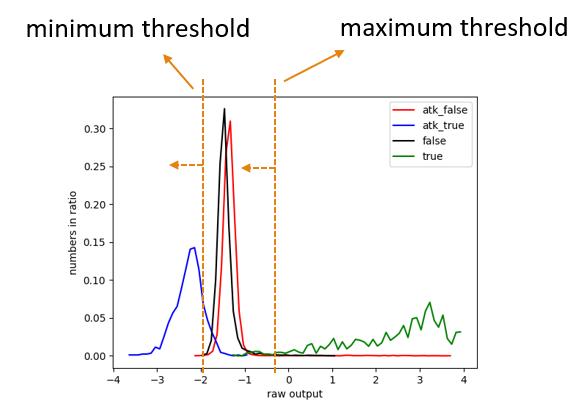}}
\caption{The histogram comparison of raw output before and after attacks. The target model is trained with the proposed loss. This figure shows that false outputs are very hard to attack (the “atk\_” curve), so the range (the two thresholds) of false raw outputs can be used to test whether the value of true output is normal.}
\label{fig:why proposed method works}
\end{center}
\vskip -0.2in
\end{figure}

After applying the proposed training loss, only the true output can be attacked. When adversarial attacks happen, the true raw output is pressed down with false outputs kept still. However, we do not know which output is the true output in the real scenario. Instead, we can only know the maximum raw output is much higher than other raw outputs in normal cases. When adversarial attacks happen, the maximum raw output will be much lower or even no longer the maximum one. In some cases, the original maximum raw output becomes the minimum one and is much lower than other raw outputs. Therefore, two thresholds are applied to check the maximum and the minimum raw output. An adversarial sample is detected when the maximum raw output is lower than the threshold or the minimum raw output is lower than the other threshold. The figure \ref{fig:why proposed method works} shows why the proposed method works. Both thresholds are selected by the range of false raw output.

\section{Experiments}
The experimental details are in the appendix, including threshold choosing, training environment, training parameter, and training tricks.

\subsection{Compared with Other Unsupervised Methods}
Although adversarial attacks are used during training, the proposed method is not a supervised method. At first, the attacks during training are not learned directly. Only true raw output is attacked in some data and only the false raw outputs of these data are included in the training loss. What's more, the applied attack type is unused in all regular attack types. The strength of this attack is also weaker than common attacks. Given $\epsilon=8$ for infinite norm and 2 for an updated step length of PGD attack, the iteration is only 5 rather than 10 in most paper settings. The most important is that the proposed detection works well with every white-box attack we've tried, and supervised detection methods can't work with unknown attacks. Therefore, we consider the proposed method as an unsupervised method.

Following the results from the survey paper of  aldahdooh \yrcite{aldahdooh2022adversarial}, the setting of adversarial attacks is a little different: the iteration of attack is 100, and there is no constraint $\epsilon$ of deep fool (DF) attack \cite{moosavi2016deepfool}. We choose infinite norm $L_{\infty}$ for constraint $\epsilon$. The test data of adversarial samples are generated by the Adversarial Robustness Toolbox \yrcite{nicolae2018adversarial}. The proposed method works extremely well against all types of  the given white box attacks \cite{goodfellow2014explaining, kurakin2018adversarial, moosavi2016deepfool, madry2017towards}. Furthermore, the iterations are 100 rather than the common setting of 10, so the attack strength is much higher than the usual setting. The proposed method works worst in DF ($\infty$) because the difference is too large (reaches 255 in some pixels) and even humans can see the difference. Under other attacks, the proposed method's true positive rate (TPR) defeats all other methods except NIC, but the false positive rate (FPR) is much lower than NIC. The result is in table \ref{unsupervised vs white box attacks table}.

\subsection{Ablation Study}
In the ablation study, AutoAttack \cite{croce2020reliable} is also included because it can evaluate the ability of defense, also called adversarial robustness,  more fairly than other attacks. The strength of attacks will follow the common setting:$\epsilon=8$ with 10 iterations and $\epsilon=16$ with 20 iterations. $\epsilon$ is based on an infinite norm, and the update rate is 2.

\subsubsection{The Effect of Multi-Hot KL Divergence for Independent-Class Training vs Cross Entropy}
\label{MHKL vs CE}
The loss for true outputs can use the common CE loss, but it makes the results worse because some true raw outputs become lower. The table \ref{MHKL vs CE table} shows that FPR is higher even with lower TPR.
\begin{table}[H]
\caption{The true positive rate (TPR\%) and false positive rate (FPR\%) of the proposed method vs CE against well-known white box attacks on CIFAR-10.}
\label{MHKL vs CE table}
\vskip 0.15in
\begin{center}
\begin{small}
\begin{sc}
\begin{tabular}{lcccc}
\toprule
& \multicolumn{2}{c}{CE} & \multicolumn{2}{c}{the proposed loss}  \\
\cmidrule(lr){2-3}\cmidrule(lr){4-5}
Attack type($\epsilon$)
&TPR&FPR&TPR&FPR\\
\midrule
PGD (8)&91.8&6.9&\textbf{94.2}&\textbf{2.5} \\
PGD (16)&85.1&6.9&\textbf{95.8}&\textbf{2.5} \\
DF (8)&99.5&6.9&\textbf{99.9}&\textbf{2.5} \\
DF (16)&93.5&6.9&\textbf{99.9}&\textbf{2.5} \\
AutoAttack (8)&79.2&6.9&\textbf{100}&\textbf{2.5} \\
AutoAttack (16)&79.4&6.9&\textbf{100}&\textbf{2.5} \\
\bottomrule
\end{tabular}
\end{sc}
\end{small}
\end{center}
\vskip -0.1in
\end{table}

\subsubsection{The Effect of Removing the One-Hot Output Trend}
When training without adversarial samples, the true raw outputs are higher in the classes with worse accuracy in the original classification task, and the accuracy of the original classification task is also higher. However, the trend of keeping one-hot outputs makes it vulnerable to any attack that presses down the true output, just like table \ref{training without adversarial samples table}. The TPR without adversarial samples is high against AutoAttack because the true outputs are lower than the minimum threshold.
\begin{table}[H]
\caption{The true positive rate (TPR\%) and false positive rate (FPR\%) of the proposed method vs training without adversarial (adv.) samples against well-known white box attacks on CIFAR-10.}
\label{training without adversarial samples table}
\vskip 0.15in
\begin{center}
\begin{small}
\begin{sc}
\begin{tabular}{lcccc}
\toprule
& \multicolumn{2}{c}{without adv. data} & \multicolumn{2}{c}{with adv. data}  \\
\cmidrule(lr){2-3}\cmidrule(lr){4-5}
Attack type($\epsilon$)
&TPR&FPR&TPR&FPR\\
\midrule
PGD (8)&23.5&0&\textbf{94.2}&\textbf{2.5} \\
DF (8)&0.2&0&\textbf{99.9}&\textbf{2.5} \\
AutoAttack (8)&99&0&\textbf{100}&\textbf{2.5} \\
\bottomrule
\end{tabular}
\end{sc}
\end{small}
\end{center}
\vskip -0.1in
\end{table}
\subsubsection{The Performance Comparison with and without the Minimum Output Threshold}
\label{effect of minimum threshold}
The minimum output threshold is an engineering method to improve the accuracy of AD. The most noteworthy part is the ability of false raw outputs to defend against adversarial attacks, and it is still not perfect now. Here show the result without minimum output threshold and the corresponding AUROC to explain the ability of defense more briefly. Attackers may adjust true outputs carefully to avoid triggering the minimum output threshold, so improving the resistance ability of the false output is still the research key point. 
\begin{table}[H]
\caption{The true positive rate (TPR\%), false positive rate (FPR\%), and AUROC of the proposed method with and without the minimum output threshold against well-known white box attacks on CIFAR-10.}
\label{Minimum Output Threshold comparison table}
\vskip 0.15in
\begin{center}
\begin{small}
\begin{sc}
\begin{tabular}{lccccc}
\toprule
& \multicolumn{3}{c}{without} & \multicolumn{2}{c}{with}  \\
\cmidrule(lr){2-4}\cmidrule(lr){5-6}
Attack type($\epsilon$)
&AUROC&TPR&FPR&TPR&FPR\\
\midrule
PGD (8)&97.8&92.1&2.5&\textbf{94.2}&2.5\\
PGD (16)&94.8&89.1&2.5&\textbf{95.8}&2.5\\
DF (8)&99.9&99.9&2.5&99.9&2.5\\
DF (16)&99.9&99.9&2.5&99.9&2.5\\
AutoAttack (8)&89.6&83.2&2.5&\textbf{100}&2.5\\
AutoAttack (16)&91.7&71&2.5&\textbf{100}&2.5\\
\bottomrule
\end{tabular}
\end{sc}
\end{small}
\end{center}
\vskip -0.1in
\end{table}

\section{Conclusions and Future Works}
We find that constraining false raw output to uniform distribution can reduce unnecessary feature learning for false outputs and help false output defend against adversarial attacks. Although keeping the model's result correct is still impossible, these adversarial samples can be easily detected by checking the maximum and the minimum raw outputs of the target model. Based on this property, We proposed an unsupervised AD method with high accuracy and much lower training time overhead than adversarial training. There is no extra model, extra training, extra computing, or difficult hyperparameter tuning. There is no accuracy trade-off for the original classification task.

What's more, the proposed training method may help in other ways. CE keeps learning accurate classes but neglects inaccurate classes, and the proposed loss for true outputs can solve this problem. Therefore, the proposed loss for true outputs may be useful when the dataset label is unbalanced or when some classes are harder to train.

 On the other hand, we still do not understand why training with some adversarial samples will press down the true raw outputs. If this problem is solved, the FPR can be even lower and the detection method will finally be perfect.
 
\section*{Acknowledgements}
This work was supported in part by the National Science and Technology Council, under Grant NSTC  110-2221-E-002-133-MY2, and Qualcomm through a Taiwan University Research Collaboration Project. 

\bibliography{refs}
\bibliographystyle{icml2023}

%%%%%%%%%%%%%%%%%%%%%%%%%%%%%%%%%%%%%%%%%%%%%%%%%%%%%%%%%%%%%%%%%%%%%%%%%%%%%%%
%%%%%%%%%%%%%%%%%%%%%%%%%%%%%%%%%%%%%%%%%%%%%%%%%%%%%%%%%%%%%%%%%%%%%%%%%%%%%%%
% APPENDIX
%%%%%%%%%%%%%%%%%%%%%%%%%%%%%%%%%%%%%%%%%%%%%%%%%%%%%%%%%%%%%%%%%%%%%%%%%%%%%%%
%%%%%%%%%%%%%%%%%%%%%%%%%%%%%%%%%%%%%%%%%%%%%%%%%%%%%%%%%%%%%%%%%%%%%%%%%%%%%%%
\newpage
\appendix
\onecolumn
\section{Experimental Details}

When some samples' predictions are wrong, their maximum output without attack will be much lower even below the threshold. Therefore, to ensure the target model's accuracy not changing the accuracy of AD, we only adopt the test data with the correct answer before attacks. 

Both thresholds for the proposed AD method are chosen based on false raw output distribution on CIFAR-10 test data. The maximum threshold is the 99th percentile, higher than 99\% false raw outputs, and the minimum threshold is the lowest false raw output.

The training process runs on the RTX 3090 based on Tensorflow. The training dataset is CIFAR-10 \cite{krizhevsky2009learning}. The model structure in the experiments is resnet18\cite{he2016deep} customized for CIFAR-10. We translate the popular PyTorch code of resnet for CIFAR-10 into Tensorflow. The training epoch is 60. The training optimizer is SGD with a momentum of 0.9, and the learning rate is 0.1 for the first 50 epochs and 0.01 for the last 10 epochs. The proposed training loss keeps the accuracy of the original classification task almost the same,  around 86\%. The ratio of attack samples in the mini-batch is 10\%. The attack samples used in training are generated by the PGD attack with $\epsilon=8$ for infinite norm, 2 for an update step length, and 5 for update step iteration given the input value from 0 to 255.

Both the training losses are applied to raw outputs rather than softmax outputs. The training procedure needs to warm up with lower weight for false output's loss, 0.2 for the first 5 epochs and 1 for the last epochs. The weight for true output's loss keeps 1 during training. Notice that the kernel weight of the last layer of the model should be initialized with 0 to avoid overflow and the dead gradient of the loss for false outputs. In this work, the bias $b$ is $\log{\frac{n_f}{n_t}}-\log{9}$ because the number of false outputs in CIFAR-10 is 9. This setting helps compare the proposed loss for true outputs and CE. It also has to mention that the softmax in $Loss_{false\_output}$ is very easy to overflow. When some values overflow, they are clipped to avoid infinite numbers or NaN values, and these outputs die because no gradient goes through them. A trick to deal with this is to initialize the kernel of the last layer with all zero or some similar ways. Besides, $Loss_{false\_output}$ is fed with $\boldsymbol{Y_F}$ and $-\boldsymbol{Y_F}$ and average both to penalize the false raw outputs with extremely high or low values at the same time.
%%%%%%%%%%%%%%%%%%%%%%%%%%%%%%%%%%%%%%%%%%%%%%%%%%%%%%%%%%%%%%%%%%%%%%%%%%%%%%%
%%%%%%%%%%%%%%%%%%%%%%%%%%%%%%%%%%%%%%%%%%%%%%%%%%%%%%%%%%%%%%%%%%%%%%%%%%%%%%%

\end{document}